\documentclass[conference]{IEEEtran}
\IEEEoverridecommandlockouts
\usepackage{amsmath,amssymb,amsfonts,graphicx,cite,multirow,subfigure}
\usepackage{algorithmic}
\usepackage{textcomp}
\def\BibTeX{{\rm B\kern-.05em{\sc i\kern-.025em b}\kern-.08em
    T\kern-.1667em\lower.7ex\hbox{E}\kern-.125emX}}

\begin{document}
\title{Feature Dimensionality Reduction for Video\\ Affect Classification: A Comparative Study}

\author{\IEEEauthorblockN{Chenfeng Guo}\\
\IEEEauthorblockA{\textit{School of Printing and Packaging}\\ \textit{Wuhan University}\\ \textit{Wuhan, China} \\ E-mail: guochenfeng@whu.edu.cn}
\and
\IEEEauthorblockN{Dongrui Wu, \textit{Senior Member, IEEE}}\\
\IEEEauthorblockA{\textit{School of Automation}\\ \textit{Huazhong University of Science and Technology}\\
\textit{Wuhan, China}  \\
E-mail: drwu@hust.edu.cn}}

\maketitle

\begin{abstract}
Affective computing has become a very important research area in human-machine interaction. However, affects are subjective, subtle, and uncertain. So, it is very difficult to obtain a large number of labeled training samples, compared with the number of possible features we could extract. Thus, dimensionality reduction is critical in affective computing. This paper presents our preliminary study on dimensionality reduction for affect classification. Five popular dimensionality reduction approaches are introduced and compared. Experiments on the DEAP dataset showed that no approach can universally outperform others, and performing classification using the raw features directly may not always be a bad choice.
\end{abstract}


\section{Introduction}

Affective computing \cite{Picard1997} is ``\emph{computing that relates to, arises from, or influences emotions.}" It is very important in human-machine interaction, as humans cannot have long-lasting intimate relationships with machines if they cannot understand our affects and respond appropriately.

Both affect classification and regression have been extensively studied in the literature \cite{drwuVRST2010,drwuInterSpeech2010,Martinez2013,drwuACS2011,drwuTL2011}. For affect classification, the most commonly used categories are the six basic emotions (anger, disgust, fear, happiness, sadness, and surprise) proposed by Ekman et al. \cite{Ekman1987}. For regression, affects are usually represented as numbers in the 2D space of arousal and valence \cite{Russell1980}, or in the 3D space of arousal, valence, and dominance \cite{Mehrabian1980}. Recently, Yannakakis et al. \cite{Yannakakis2017} also argued that the nature of emotions is ordinal, and hence preference learning \cite{Yannakakis2009} should also play an important role in affective computing.

Various input signals could be used in affective computing, e.g., speech \cite{Lee2005,drwuICME2010}, facial expressions \cite{Pantic2000, Fasel2003}, physiological signals \cite{Fairclough2009, drwuVRST2010}, and multimodal combination \cite{Metallinou2010, Zeng2009}. Numerous features could be extracted from each modality. For example,  6,373 acoustic features were extracted by OpenSMILE \cite{Eyben2010b} in the InterSpeech 2013 Computational Paralinguistics Challenge. 465 Riemannian tangent space features were extracted from 30-channel EEG signals in \cite{drwuRG2017}. The number would increase to 2,080 for 64-channel EEG signals, and 8,256 for 128-channel. And, 22,881 features were extracted from 64-channel EEG signals in \cite{Jenke2014} for emotion recognition.

On the contrary, affects are very subjective, subtle, and uncertain. So, usually multiple human assessors are needed to obtain the groundtruth affect label for each video, audio, or facial expression. So, generally it is not easy to obtain a large number of labeled training samples in affective computing. As a result, the curse of dimensionality \cite{Hastie2009} becomes very significant in affect recognition, which implies high computational cost and poor generalization performance. Thus, it is critical to perform dimensionality reduction in affective computing. Though lots of dimensionality reduction approaches have been proposed in the literature \cite{Chandrashekar2014,Li2016}, to the authors' knowledge, they have not been extensively studied specifically for affect recognition.

This paper compares five representative dimensionality reduction approaches in video affect classification. It represents our preliminary study of a comprehensive investigation on dimensionality reduction for affective computing.

The remainder of this paper is organized as follows: Section~\ref{sect:DR} introduces five representative dimensionality reduction approaches used in our study. Section~\ref{sect:dataset} describes the DEAP dataset, the raw visual and audio features, and the experimental results. Section~\ref{sect:conclusions} draws conclusions and points out several future research directions.

\section{Dimensionality Reduction} \label{sect:DR}

Dimensionality reduction approaches can be categorized into two main classes \cite{Li2016}: feature extraction and feature selection.

Feature extraction projects a high-dimensional feature space to a low-dimensional one, which is usually a linear or nonlinear combination of the original feature space. Typical feature extraction approaches include Principal Component Analysis (PCA) \cite{Jolliffe2002}, Linear Discriminant Analysis (LDA) \cite{Mika1999}, Canonical Correlation Analysis (CCA) \cite{Hardoon2004}, Singular Value Decomposition \cite{Golub2012}, Locally Linear Embedding (LLE) \cite{Roweis2000}, etc.

Feature selection directly selects a subset of relevant features to be used in machine learning. According to how feature selection is integrated with the machine learning model, feature selection approaches can be categorized into three groups \cite{Chandrashekar2014}:
\begin{enumerate}
\item \emph{Filter methods}, which select the features independent of the machine learning model. Typical criteria used in filter methods include correlation, mutual information \cite{Vergara2015}, Relief \cite{Kira1992}, etc.
\item \emph{Wrapper methods}, which wrap the machine learning model into a search algorithm to find the optimal feature set that gives the highest learning performance. Typical wrapper methods include sequential forward/backward selection \cite{Pudil1994}, and heuristic search algorithms (e.g., genetic algorithms \cite{Godlberg1989}, particle swarm optimization \cite{Kennedy1995}, etc) for feature subset selection.
\item \emph{Embedded methods}, which include feature selection as part of the machine learning model training process. Typical embedded methods include LASSO \cite{Tibshirani1996}, 1-norm support vector machines (SVMs) \cite{Zhu2004a}, etc.
\end{enumerate}

Five representative dimensionality reduction approaches are introduced in more details next.

\subsection{Principal Component Analysis (PCA)}

PCA \cite{Jolliffe2002} uses an orthogonal transformation to convert a set of observations into a set of values on the principal components, which are linearly uncorrelated and ordered so that the first few retain most of the variation present in all of the original variables.

Let $\mathbf{X}\in \mathbb{R}^{N\times d}$ be the data matrix with $N$ observations of $d$ dimensions. Assume $\mathbf{X}$ has been pre-processed such that each column has mean zero. Then, its principal components decomposition is:
\begin{align}
\mathbf{X}'=\mathbf{XW}
\end{align}
where the columns of $\mathbf{W}\in \mathbb{R}^{d\times d}$ are the eigenvectors of $\mathbf{X}^T\mathbf{X}$, sorted in descending order according to the corresponding eigenvalues. $\mathbf{X}'$ can then be used to represent $\mathbf{X}$ in the new space spanned by the columns of $\mathbf{W}$. Usually the most variation in $\mathbf{X}$ is distributed along the first $m$ columns of $\mathbf{W}$, and hence only the first $m$ ($m<d$) columns of $\mathbf{X}'$ are enough to represent $\mathbf{X}$, i.e., the dimensionality can be effectively reduced from $d$ to $m$.

There can be different approaches to determine $m$. One is to find the minimum number of eigenvalues of $\mathbf{X}^T\mathbf{X}$ such that their sum exceeds a pre-defined threshold, e.g., 95\% of the sum of all eigenvalues. In this paper we used 5-fold cross-validation to find the $m$ that gave the highest classification accuracy on the training data.

\subsection{Sequential Forward Selection (SFS)}

Sequential forward selection (SFS) \cite{Pudil1994} is a very common and intuitive feature selection approach, in which features are sequentially added until the addition of further features does not improve the cross-validation performance. Starting from an empty feature set, SFS creates candidate feature subsets by successively adding each of the features not yet selected. For each candidate feature subset, SFS performs cross-validation to determine the optimal one.

\subsection{ReliefF}

The original Relief algorithm was proposed by Kira and Rendell in 1992 \cite{Kira1992}. ReliefF \cite{Kononenko1994,Robnik2003} is its improved version.

Relief is an iterative procedure that ranks the importance of the features according to how well their values distinguish between their nearest neighbors in different classes. For binary classification, in each iteration Relief first randomly selects a training sample $\mathbf{x}_i$, and identifies its two nearest neighbors, one from each class. Denote the neighbor from the same class (called \emph{nearest hit}) as $\mathbf{h}$, and the one from the other class (called \emph{nearest miss}) as $\mathbf{m}$. Then, Relief updates the feature weight vector $\mathbf{w}=[w_1,...,w_d]^T$ as:
\begin{align}
w_j=w_j-\frac{diff(\mathbf{x}_{ij},\mathbf{h}_j)}{M}+\frac{diff(\mathbf{x}_{ij},\mathbf{m}_j)}{M}, \quad j=1,...,d
\end{align}
where $M$ is the pre-defined number of iterations, and $diff(\mathbf{x}_{ij},\mathbf{h}_j)$ is  the difference between the $j$th feature for $\mathbf{x}_{ij}$ and $\mathbf{h}_j$. When the feature is discrete/categorical,
\begin{align}
diff(\mathbf{x}_{ij},\mathbf{h}_j)=\left\{\begin{array}{cc}
                                            1, & \mathbf{x}_{ij}\neq\mathbf{h}_j \\
                                            0, & \mbox{otherwise}
                                          \end{array}\right. \label{eq:Relief}
\end{align}
When the feature is continuous,
\begin{align}
diff(\mathbf{x}_{ij},\mathbf{h}_j)=\mathbf{x}_{ij}-\mathbf{h}_j
\end{align}
provided that the continuous feature has been normalized to $[0,1]$. $diff(\mathbf{x}_{ij},\mathbf{m}_j)$ is computed similarly.

In summary, the rationale of (\ref{eq:Relief}) is to penalize features that have different values for neighbors from the same class (i.e., features that may lead to wrong classification), and reward features that have different values for neighbors from different classes (i.e., features with good distinguishibility).

The ReliefF algorithm \cite{Kononenko1994,Robnik2003} improved Relief from the following three perspectives:
\begin{enumerate}
\item It is more robust, by using $k$ nearest neighbors from each class, instead of only one.
\item It can handle multi-class classification instead of binary classification only.
\item It can deal with incomplete and noisy data.
\end{enumerate}
In this paper we used $k=10$ in ReliefF, as suggested in \cite{Kononenko1994}.

Because ReliefF ranks the features instead of selecting a subset of them, in this paper we applied PCA to the $d/2$ most important raw features, and then used 5-fold cross-validation on the training data to find the optimal number of PCA features.

\subsection{Minimal-Redundancy-Maximal-Relevance (mRMR)}

For discrete/categorical variables $X$ and $Y$, their mutual information $I(X;Y)$ is defined as:
\begin{align}
I(X;Y)=\sum_{y\in Y}\sum_{x\in X}p(x,y)\log\frac{p(x,y)}{p(x)p(y)}
\end{align}
where $p(x,y)$ is the joint probabilistic distribution, and $p(x)$ and $p(y)$ are the marginal probabilities.

Consider a $C$-class classification problem, and we want to select a feature subset $S$ with $m$ discrete features $\{x_i\}_{i=1}^m$. The minimal-redundancy-maximal-relevance (mRMR) \cite{Peng2005,Ding2003} feature selection approach optimizes the following mutual information difference criterion:
\begin{align}
\max D-R \label{eq:mid}
\end{align}
or the mutual information quotient criterion:
\begin{align}
\max D/R \label{eq:miq}
\end{align}
where
\begin{align}
D=\frac{1}{|S|}\sum_{x_j\in S}I(x_j;c)
\end{align}
is the mean of all mutual information values between individual feature $x_j$ and the class label, and
\begin{align}
R=\frac{1}{|S|^2}\sum_{x_j,x_r\in S}I(x_j;x_r)
\end{align}
is the redundancy among all features in $S$. In practice incremental search is used to find the near-optimal feature subset $S$ \cite{Peng2005,Ding2003}.

Note that we only introduce the mRMR approach for discrete features. mRMR approaches for continuous features have also been proposed \cite{Ding2003}, but it was found that the discrete versions usually work better, so the discrete mRMR is preferred. In this paper we converted each continuous feature $x_j$ into three discrete levels, by thresholds $\bar{x}_j\pm std(x_j)$ (i.e., values larger than $\bar{x}_j+std(x_j)$ were mapped to $1$, smaller than $\bar{x}_j-std(x_j)$ to $-1$, and the rest to $0$), where $\bar{x}_j$ is the mean of $x_j$, and $std(x_j)$ is the standard deviation. The objective function (\ref{eq:miq}) was used in our study, because it gave slightly better performance than (\ref{eq:mid}), as demonstrated in \cite{Ding2003}. Finally, the optimal number of features $m$ was determined by 5-fold cross-validation on the training dataset, from the value set of $\{1,...,d/2\}$.

\subsection{Neighborhood Component Analysis (NCA)}

Neighborhood component analysis (NCA) \cite{Yang2012} is a non-parametric feature selection scheme that can be used for both classification and regression. Next we introduce its formulation for classification.

Let $\{(\mathbf{x}_i,y_i)\}_{i=1}^N$ be $N$ training samples, where $\mathbf{x}_i\in \mathbb{R}^d$ is the feature vector and $y_i\in\{1,...,C\}$ is the corresponding class label. In classification, NCA tries to find a feature weighting vector $\mathbf{w}=[w_1,...,w_d]^T$ that maximizes the average leave-one-out  cross-validation (LOOCV) accuracy.

The weighted distance between $\mathbf{x}_i$ and $\mathbf{x}_l$ is:
\begin{align}
d(\mathbf{x}_i,\mathbf{x}_l)=\sum_{j=1}^d w_j^2|x_{ij}-x_{lj}|
\end{align}
Consider a nearest neighbor classifier, which randomly selects the neighbor $\mathbf{x}_l$ for $\mathbf{x}_i$ according to the following probabilities:
\begin{align}
p_{il}=\left\{\begin{array}{cl}
         \frac{\kappa(d(\mathbf{x}_i,\mathbf{x}_l))}
         {\sum_{n\neq l}\kappa(d(\mathbf{x}_i,\mathbf{x}_n))}, & i\neq l \\
         0, & i=l
       \end{array}\right.
\end{align}
where $\kappa(d(\mathbf{x}_i,\mathbf{x}_l))=\exp(-d(\mathbf{x}_i,\mathbf{x}_l)/\sigma)$ is a kernel function, in which $\sigma$ is the kernel width. Then, the LOOCV classification accuracy for $\mathbf{x}_i$ is:
\begin{align}
p_i=\sum_{l\neq i} p_{il}y_{il}
\end{align}
where
\begin{align}
y_{il}=\left\{\begin{array}{cc}
                1, & y_i=y_l \\
                0, & y_i\neq y_l
              \end{array}\right.
\end{align}

NCA for classification then maximizes the following regularized objective function:
\begin{align}
f(\mathbf{w})=\sum_{i=1}^N p_i -\lambda\sum_{j=1}^d w_j^2
\end{align}
where $\lambda$ is a regularization parameter. $\sigma=1$ and $\lambda=1$ were used in this paper. After finding $\mathbf{w}$, we sorted $w_j$ in descending order, identified the first a few such that their sum accounts for at least 95\% of $\sum_{j=1}^d w_j$, and selected the corresponding features in classification.

\section{Experiments} \label{sect:dataset}

This section describes the DEAP dataset used in our study, the raw visual and audio features, and the experimental results

\subsection{The DEAP Dataset} \label{sect:DEAP}

The DEAP dataset \cite{Koelstra2012} was used in our study. It consists of 40 1-minute music video clips, each of which had been evaluated by 14-16 assessors online. Each assessor watched the music videos and rated them on a discrete 9-point scale for valence, arousal, and dominance. Among the 40 videos, 10 had high arousal and high valence, 10 high arousal and low valence, 10 low arousal and high valence, and 10 low arousal and low valence. We would like to classify valence, arousal and dominance independently into two levels (high and low), from the visual, audio, and video signals.

\subsection{Visual Features}

The visuals were first converted to MPEG format files at 25 FPS. The 16 features extracted, shown in Table~\ref{tab:visual}, consisted of the following valence-related frame-based static features and arousal-related motion features:
\begin{enumerate}
\item \textbf{Static features}: We used the lighting key \cite{Yazdani2013}, lightness, and color variance to describe the brightness and color information of frames. The lighting key was defined as the product of the mean and variance of the V-channel in the HSV color space. The lightness was defined as the maximum, minimum, and mean value of the V-channel in the HSV color space. To calculate the color variance, the key-frames were first identified by comparing the histogram distances of two adjacent frames, and then the mean of the determinant of the covariance matrix of the L, U, and V components in the CIELUV color space of the key-frames were computed.

\item \textbf{Motion features}: Motion features show the changes between frames and the movements of shot, including the shot change rate, shot length, visual excitement, and motion component. The shot change rate was simply defined as the numbers of key-frames. The shot length features consisted of the longest, shortest, and mean shot lengths. The visual excitement, which measures the degree of video arousal, was calculated from the amount of local pixel changes according to the definition in \cite{Wang2006c}. The motion components were calculated by accumulating the absolute values of the \emph{x} and \emph{y} of the motion vectors, and their sum of squares.
\end{enumerate}

\begin{table}[htbp] \centering \setlength{\tabcolsep}{2mm}
\caption{The 16 visual features.} \label{tab:visual}
\begin{tabular}{|c|c|c|} \hline
\textbf{Feature category}&\textbf{Number}&\textbf{Value} \\ \hline
Lighting key&	3&	Mean \\ \hline
Lightness&	3&	Median \\ \hline
Color variance&	1&	Mean\\ \hline
Shot change rate&	1&	Mean\\ \hline
Shot length&	3&	Mean\\ \hline
Visual excitement&	2&	Mean, variance\\ \hline
Motion component&	3&	Mean\\ \hline
\end{tabular}
\end{table}

\subsection{Audio Features}

Mono MP3 format audio was first extracted from each video at a sampling rate of 44.1 kHz. Each audio was spilt into frames and frame-level features were extracted, as follows:
\begin{enumerate}
\item \textbf{Low-level features}, which describe the basic properties of audio in time- and frequency- domains, including the spectral centroid, band energy radio, delta spectrum magnitude, zero crossing rate, short-time average energy, and pitch. More details about these low-level features can be found in \cite{Li2001}.
\item \textbf{Silence ratio}, which is the ratio of the amount of silence frames to the time window \cite{Chen2006}. A frame is considered as a silence frame when its root mean square is less than 50\% of the mean root mean square of the fixed-length audio fragments.
\item \textbf{MFCCs and LPCCs}. In order to combine the static and dynamic characteristics of audio signals, 12 Mel Frequency Cepstral Coefficients (MFCCs), 11 Linear Predictive Cepstral Coefficients (LPCCs), and 12 first-order differential MFCC coefficients were calculated.
\item \textbf{Formant}, which reflects the resonant frequencies of the vocal tract. Formant frequencies F1-F5 in each frame were extracted.
\end{enumerate}
We then computed the mean and/or variance of these frame-level features, resulting in a total of 76 audio features, as shown in Table~\ref{tab:audio}.

\begin{table}[htbp]\centering \setlength{\tabcolsep}{2mm}
\newcommand{\tabincell}[2]{\begin{tabular}{@{}#1@{}}#2\end{tabular}}
\caption{The 76 audio features.} \label{tab:audio}
\begin{tabular}{|c|c|c|} \hline
\textbf{Feature category}&\textbf{Number}&\textbf{Value} \\ \hline
\tabincell{c}{Spectral centroid,\\Band energy radio,\\Delta spectrum magnitude,\\Zero crossing rate, \\Pitch,\\Short-time average energy }& 12 & Mean, variance \\  \hline
Silence ratio &  1  &  Mean \\ \hline
\tabincell{c}{MFCC coefficients,\\Delta MFCC,\\LPCC}&\tabincell{c}{24\\12\\22}&\tabincell{c}{Mean, variance\\Mean\\Mean, variance} \\ \hline
Formant &5&	Mean \\  \hline
\end{tabular}
\end{table}

\subsection{Experimental Results} \label{sect:experiments}

We compared the performances of the five dimensionality reduction approaches introduced in Section~\ref{sect:DR}, plus a baseline approach (denoted as \emph{Raw}) that does not use any dimensionality reduction, i.e., all the extracted raw features were used. The performance measure was the LOOCV accuracy on the 40 videos.

\begin{figure}[htbp]\centering
\subfigure[]{\includegraphics[width=\linewidth,clip]{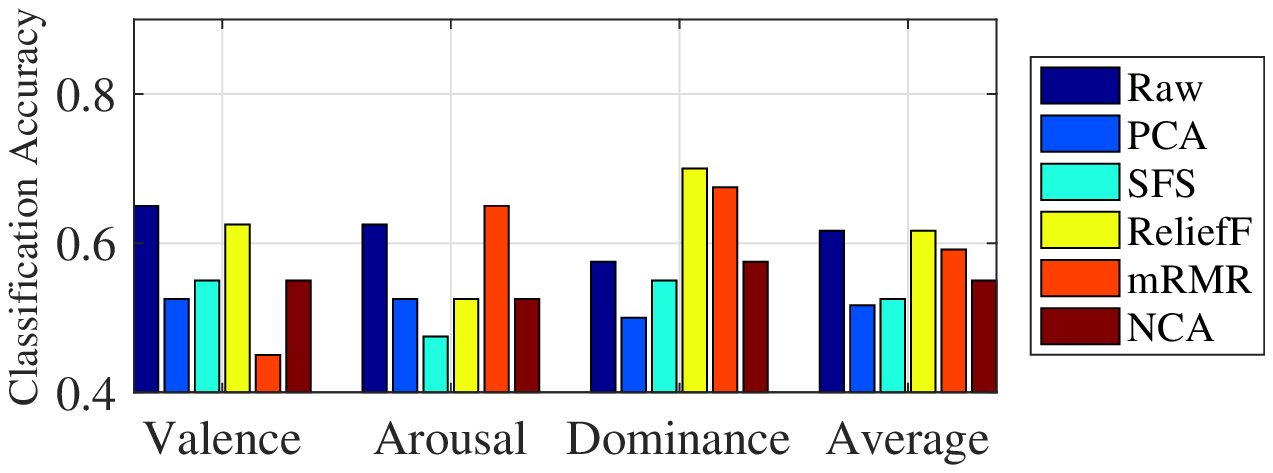}} \label{fig:visual}
\subfigure[]{\includegraphics[width=\linewidth,clip]{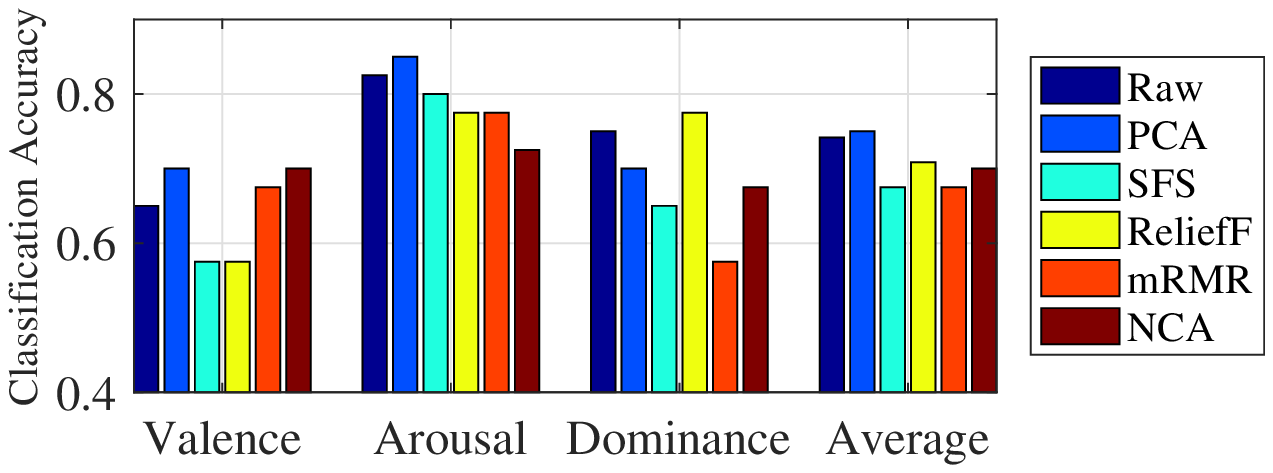}} \label{fig:audio}
\subfigure[]{\includegraphics[width=\linewidth,clip]{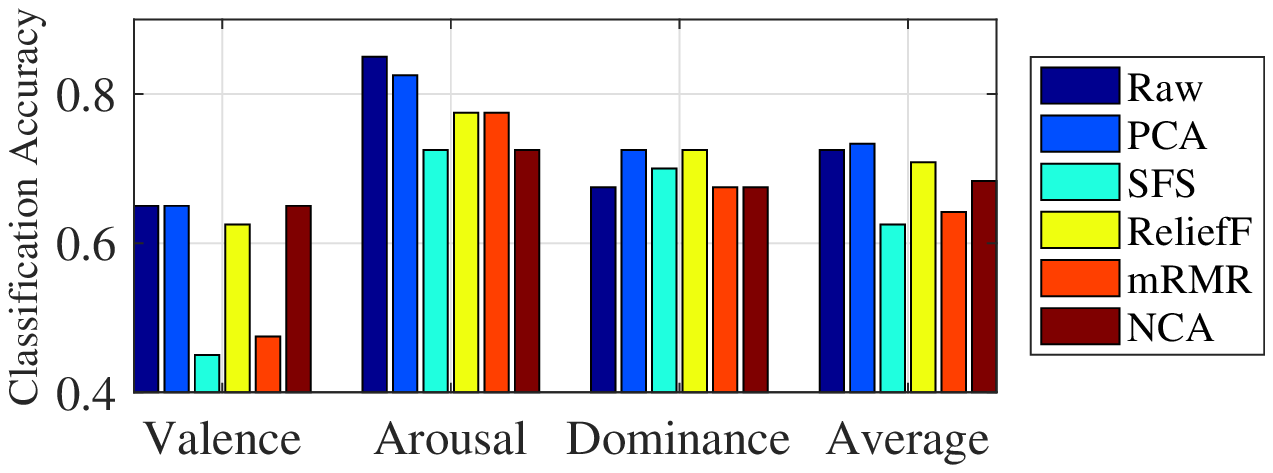}} \label{fig:video}
\caption{LOOCV classification accuracies on the DEAP dataset. (a) visual only; (b) audio only; (c) video (visual + audio).}\label{fig:offline}
\end{figure}

We assume that the features for the 40 videos are all available, the labels for 39 videos are known, and we would like to estimate the label for the remaining video. We first normalized each dimension of the features to [0,1], and then applied different dimensionality reduction approaches. Finally, an radial basis function (RBF) SVM was used as the classifier, where the best SVM parameters ware found through 5-fold cross-validation on the training data (39 videos). The final LOOCV classification accuracies are shown in Table~\ref{tab:offline} (the highest ones are in bold), and also illustrated in Fig.~\ref{fig:offline}, for different feature sets. Observe that:
\begin{enumerate}
\item Generally better classification accuracies were obtained from audio than from visual. This may be because more audio features were extracted.
\item Interestingly, combining visual and audio features and then performing dimensionality reduction did not necessarily improve the classification performance. In fact, most of the time the performance was actually decreased. These results suggested that the feature selection approaches were not always able to select the global optimal features: otherwise the classification accuracies on the video features would not be lower than those on the visual or audio features alone.
\item The highest performances on different affect dimensions and different modalities were achieved by different dimensionality reduction approaches (and sometimes the raw features), and there was not a single approach that was always better than others. This is consistent with the well-known \emph{no free lunch theorems for optimization} \cite{Wolpert1997}, which state that ``\emph{for any algorithm, any elevated performance over one class of problems is offset by performance over another class}."
\item Surprisingly, on average the raw features achieved the best overall performance in our experiments. This may be because the dimensionality of our features was not high enough (the video features had 76+16=92 dimensions). In the future we will increase the dimensionality of the features, and also take the computational cost into consideration.
\item Our preliminary study showed that ReliefF and PCA were two of the better dimensionality reduction approaches among the five. More extensive comparisons will be performed in the future to verify this.
\end{enumerate}

\begin{table}[htbp] \centering \setlength{\tabcolsep}{1.5mm}
\caption{LOOCV classification accuracies of different feature selection approaches.}   \label{tab:offline}
\begin{tabular}{c|c|cccccc}   \hline
Modality & Affect & Raw & PCA & SFS & ReliefF & mRMR & NCA \\ \hline
&Valence&\textbf{0.650}&0.525&0.550&0.625&0.450&0.550\\
Visual&Arousal&0.625&0.525&0.475&0.525&\textbf{0.650}&0.525\\
&Dominance&0.575&0.500&0.550&\textbf{0.700}&0.675&0.575\\
& Average & \textbf{0.617}&0.517&0.525&\textbf{0.617}&0.592&0.550 \\ \hline
&Valence&0.650&\textbf{0.700}&0.575&0.575&0.675&\textbf{0.700}\\
Audio&Arousal&0.825&\textbf{0.850}&0.800&0.775&0.775&0.725\\
&Dominance&0.750&0.700&0.650&\textbf{0.775}&0.575&0.675\\
&Average &0.742&\textbf{0.750}&0.675&0.708&0.675&0.700\\     \hline
&Valence&\textbf{0.650}&\textbf{0.650}&0.450&0.625&0.475&\textbf{0.650}\\
Video&Arousal&\textbf{0.850}&0.825&0.750&0.775&0.750&0.725\\
&Dominance&0.675&\textbf{0.725}&0.700&\textbf{0.725}&0.675&0.675\\
&Average&0.725&\textbf{0.733}&0.625&0.708&0.642&0.683\\  \hline
\multicolumn{2}{c|}{Overall Average}&\textbf{0.694}&0.667&0.608&0.678&0.636&0.644 \\ \hline
\end{tabular}
\end{table}

%
%
%

\section{Conclusions and Future Works} \label{sect:conclusions}

Affective computing problems typically have a small number of training samples, compared with the number of possible features we could extract. Thus, dimensionality reduction becomes a necessity. This paper reports our preliminary results on dimensionality reduction for affect classification. Five popular dimensionality reduction approaches have been introduced and compared. Experiments on the DEAP dataset showed that no approach can universally outperform others, and performing classification using the raw features directly (without dimensionality reduction) may sometimes result in even better performance.

Our current study has some limitations, e.g., we only considered one dataset, and the dimensionality of the features was not high enough. We will deal with them in our future research, by considering more affective computing datasets, e.g., MAHNOB-HCI \cite{Soleymani2012}, MSP-IMPROV \cite{Busso2017}, and AMIGOS \cite{Miranda2017}, and by extracting more features, e.g., through OpenSMILE \cite{Eyben2010b}. Additionally, we will:
\begin{enumerate}
\item Optimize the parameters in the feature selection approaches, e.g., $k$ (the number of nearest neighbors) in ReliefF, the thresholds for discretization in mRMR, and $\sigma$ (the kernel width) and $\lambda$ (the regularization parameter) in NCA.

\item Investigate multi-view feature selection approaches. Because visual and audio, and sometimes also physiological signals, represent different facets of the same affect, it is more intuitive to perform feature selection in a multi-view setting, instead of combining features from different modalities directly and then performing an overall feature selection. Potential multi-view feature selection approaches include sparse group LASSO \cite{Friedman2010}, adaptive unsupervised multi-view feature selection \cite{Feng2012}, unsupervised multi-view feature selection \cite{Tang2013}, and multi-view clustering and feature learning via structured sparsity \cite{Wang2013a}.

\item Compare also the computational cost of different dimensionality reduction approaches.

\item Study also dimensionality reduction in affect regression \cite{drwuICME2010} and ranking \cite{Yannakakis2017}.
\end{enumerate}


\end{document}